\documentclass[lettersize,conference]{IEEEtran}
\IEEEoverridecommandlockouts
\usepackage{cite}
\usepackage{amsmath,amssymb,amsfonts}
\usepackage{algorithmic}
\usepackage{graphicx}
\usepackage{textcomp}
\usepackage{xcolor}
\usepackage{graphicx}
\usepackage{caption}
\usepackage{float}
\usepackage{subcaption}
\usepackage{booktabs}
\usepackage{pgfplots}
\pgfplotsset{compat=newest}
\usepackage{multirow}
\usepackage{tabularx}
\usepackage{comment}
\newcommand{\RomanNumeralCaps}[1]
{\MakeUppercase{\romannumeral #1}}

\def\BibTeX{{\rm B\kern-.05em{\sc i\kern-.025em b}\kern-.08em
    T\kern-.1667em\lower.7ex\hbox{E}\kern-.125emX}}
\begin{document}

\title{On-device Filtering of Social Media Images for Efficient Storage}

\makeatletter
\newcommand{\linebreakand}{%
  \end{@IEEEauthorhalign}
  \hfill\mbox{}\par
  \mbox{}\hfill\begin{@IEEEauthorhalign}
}
\makeatother

\author{
  \IEEEauthorblockN{Dhruval Jain}
  \IEEEauthorblockA{\textit{On-device AI} \\
    Samsung R\&D Institute\\
Bengaluru, India \\
dhruval.jain@samsung.com}
  \and
  \IEEEauthorblockN{Debi Prasanna Mohanty}
  \IEEEauthorblockA{\textit{On-device AI} \\
    Samsung R\&D Institute\\
Bengaluru, India \\
debi.m@samsung.com}
  \and
  \IEEEauthorblockN{Sanjeev Roy}
  \IEEEauthorblockA{\textit{On-device AI} \\
Samsung R\&D Institute\\
Bengaluru, India \\
sanjeev.roy@samsung.com}
  \linebreakand 
  \IEEEauthorblockN{Naresh Purre}
  \IEEEauthorblockA{\textit{On-device AI} \\
Samsung R\&D Institute\\
Bengaluru, India \\
naresh.purre@samsung.com}
  \and
  \IEEEauthorblockN{Sukumar Moharana}
  \IEEEauthorblockA{\textit{On-device AI} \\
Samsung R\&D Institute\\
Bengaluru, India \\
msukumar@samsung.com}
}

\maketitle

\begin{abstract}
Artificially crafted images such as memes, seasonal greetings, etc are flooding the social media platforms today. These eventually start occupying a lot of internal memory of smartphones and it gets cumbersome for the user to go through hundreds of images and delete these synthetic images. To address this, we propose a novel method based on Convolutional Neural Networks (CNNs) for the on-device filtering of social media images by classifying these synthetic images and allowing the user to delete them in one go. The custom model uses depthwise separable convolution layers to achieve low inference time on smartphones. We have done an extensive evaluation of our model on various camera image datasets to cover most aspects of images captured by a camera. Various sorts of synthetic social media images have also been tested. The proposed solution achieves an accuracy of 98.25\% on the Places-365 dataset and 95.81\% on the Synthetic image dataset that we have prepared containing 30K instances.

\end{abstract}


\section{Introduction}
In recent years, there has been a huge flux of data on the internet. Social media platforms have contributed significantly in increasing the volume of images circulated. Images are being used to communicate opinions on trending news through memes. According to a survey, 3.2 billion images are shared each day over the internet. A large proportion of these images are wholly synthetic and become irrelevant to the user in a short period.  
Images captured from the camera have very few sharp edge transitions and are characterized by sensor pattern noise. Jan Lukas et al. \cite{lukas} proposed the identification of digital cameras based on the sensor’s pattern noise by using a correlation detector to investigate the presence of the reference noise pattern in the given image. Corripio et al. \cite{corripio} computed wavelet features for smartphone camera identification. Artificially crafted memes or seasonal greetings have sharp edge transitions. If the image is entirely artificially generated, it doesn't have noise in its raw form. Some noise is added by the lossy compression techniques used by social media platforms. These synthetic images are generally created by adding artificial text on the camera image or stacking multiple camera images together. In this process, certain regions of the image get sharper edge transitions and uniform pixel intensities.\par
	Convolutional Neural networks can learn complex latent features based on image content by which they can distinguish between images portraying different objects or scenes. This has led to their widespread success in the object recognition task. We show that CNNs can perform equally well in distinguishing between two images that portray the same object or scene but one is either synthetically generated or has added artificial characteristics in the form of text, clip-art or image croppings. Depthwise separable convolutions significantly reduce the number of parameters to build light weight deep neural network architectures. MobileNet \cite{mobilenet} uses $3\times3$ depthwise separable convolutions which need 8 to 9 times lesser computations than standard convolutions. MobileNet-224\cite{mobilenet} used 4.2 million parameters and suffered a drop in accuracy of only 0.9\% on the Imagenet classification task compared to VGG16\cite{vgg16} which used 138 million parameters. 

This paper is divided into six sections. We have built our custom deep learning architecture whose details are discussed in section \ref{sec:Methodology}. Section \RomanNumeralCaps{4} provides an extensive evaluation of the proposed model along with visual representations illustrating the learning of the model. 

\begin{figure*}[t!]
  \hspace*{\fill}
    \begin{subfigure}[b]{0.35\textwidth}
        \includegraphics[width=\textwidth]{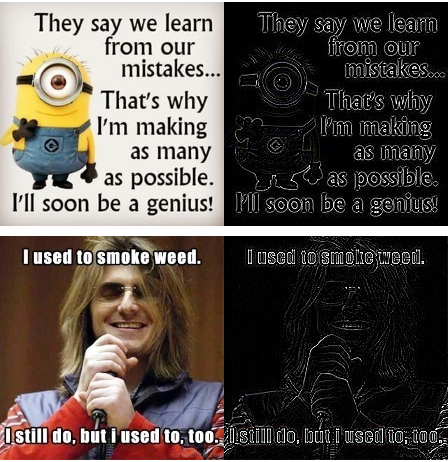}
        \caption{Synthetic Images}
    \end{subfigure}
    \hspace*{\fill}
    \begin{subfigure}[b]{0.35\textwidth}
        \includegraphics[width=\textwidth]{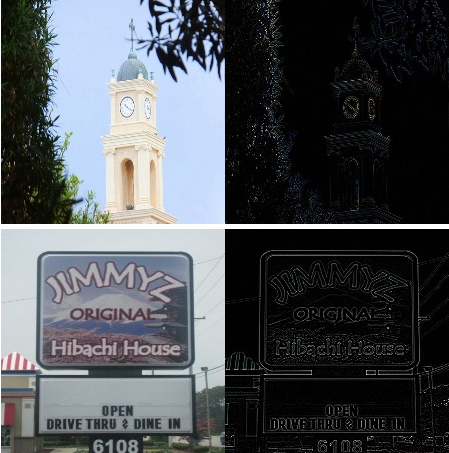}
        \caption{Camera Images}
    \end{subfigure}
	\hspace*{\fill}
    \caption{Edge Residuals}\label{fig:edges}
\end{figure*}

\section{Related Work}
Previous works like \cite{7params} and \cite{npic2} computed hand-crafted features based on color and space correlation of pixels for classification. It assumed synthetic images to have more colors than natural images as they tend to have large uniform regions of the same color. These assumptions do not hold good with artificially manipulated camera images like memes that are very popular on social media today. Due to the noise added by the lossy JPEG compression, uniform color regions may have varying pixel intensities. In \cite{7params}, eight features were computed including saturation average, SGLD histogram \cite{20}, etc and various classifiers like AdaBoost, SVMs and neural networks were tested. These feature values are averaged globally across the image, thereby failing to describe the image locally. Some camera images may have very sharp edges distributed locally in certain regions and the globally averaged feature values may fall near to those of synthetic images. \cite{7params} and \cite{npic2} also highlighted that camera images have faded edges, unlike the synthetic ones. But with advanced camera sensors that we have today, a handful of these features do not suffice and may lead to misleading results.

Classification of images based on image content has been an area of active research with the advent of deep learning. But the classification of images based on statistical properties has not received much attention. V Andrearczyk et al. \cite{texture} introduced Texture CNN. In their work, they proposed that features extracted by the fully connected layers of CNN architectures like AlexNet \cite{alexnet}, based on shape information , are of very little importance in texture understanding. Wavelet CNNs proposed by S Fujieda et al. \cite{wavelet} incorporate spectral information to enhance texture classification.

Bayar et al. \cite{bayar} proposed a CNN architecture for detecting image manipulation that suppresses the image content to learn features based on tampering. The scope of manipulation was only restricted to gaussian blurring, resampling using bilinear interpolation, median filtering, and additive white gaussian noise. They did not consider adding artificial text or image-croppings. 

 Other works in this domain include \cite{yao}, \cite{cvpr} and \cite{rahmouni} which focus on distinguishing photorealistic computer graphics from camera images. The scope of our problem is entirely different from them as we focus on artificially crafted social media images that become junk to the user in a short period of time. To best of our knowledge, we are the first to propose a lightweight CNN based architecture to perform this task on smartphones with a low inference time.

\begin{figure*}[t!]
\includegraphics[scale=0.9]{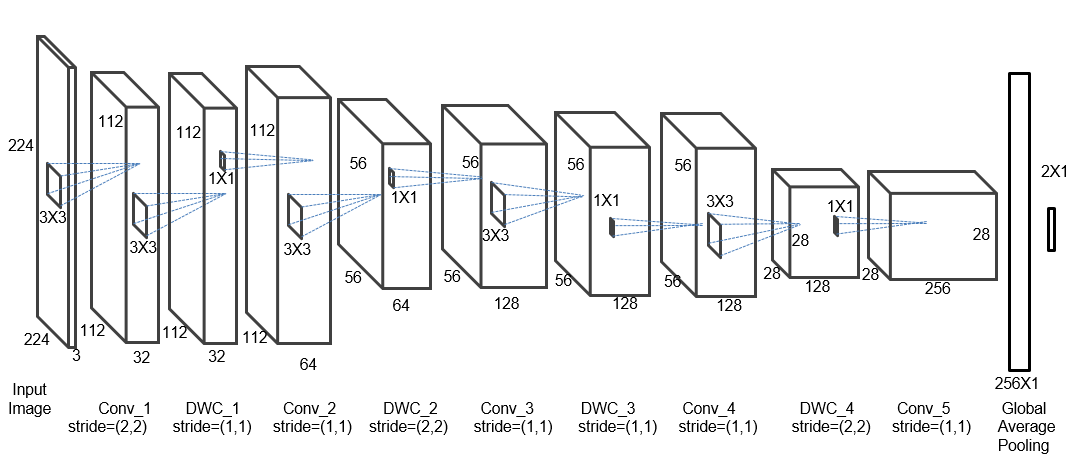}
\caption{Model Architecture for DWS\_1 }
\end{figure*}

\section{Methodology}\label{sec:Methodology}
We begin by examining the edge residuals of the camera and synthetic images. High pass filters designed by Fridrich et al. \cite{hpf} suppress the low-frequency components that represent the image content. These filters, when applied to synthetic images, are expected to produce strong edge residuals. Camera images which have very few sharp edge transitions compared to synthetic images tend to have weaker edges residuals. Fig1 (a) shows how artificial text is demarcated from the background, whereas in Fig1 (b), the edge residuals of camera images are weak even in case of the scenic text. These edge residuals must be analyzed locally, and therefore we use CNNs to learn latent features to enhance classification. The depthwise separable convolution is composed of two layers, a depthwise convolution layer followed by a pointwise convolution layer. In depthwise convolutions, unlike the standard convolutions, a single filter is applied per input channel. In pointwise convolutions, $1\times1$ kernels are used to compute the linear combination of the output of the depthwise convolution layer. The m$^{th}$ channel of output feature map $\hat{\textbf{G}}$ is given by 
\begin{equation} 
\hat{\textbf{G}}_{k,\ l,\  m} \ =\ \sum _{i,j}\hat{\textbf{K}}_{i,\\j,\\m} \ \cdotp \ \textbf{F}_{k+i-1,\ l+j-1,\ m}
\end{equation}
where $\hat{\textbf{K}}$ is the depthwise convolution kernel which is applied to the  m$^{th}$ channel of the input \textbf{F}.

Images are downsampled to $224\times 224$ before they are fed into the model to achieve faster inference on-device. Even though this results in a significant loss of information, our model is able to approximate the target function well and produce good results on the standard datasets as shown in Table \ref{sample-table}. No other preprocessing is required. We have come up with three different architectures to test our hypothesis. They are described as follows.

\begin{itemize}
\item Fig2. shows our model architecture which we call \textbf{DWS\_1}. Input image has dimensions $224\times 224\times3$. Each block shown in the figure is the output of the previous layer. Conv\_i layers, where \textit{i = 1, 2 . . 5} are standard convolution layers and DWC\_i, where \textit{i = 1, 2, 3, 4} are depthwise convolution layers. Conv\_1 layer has 32 filters and performs $3\times3$ convolutions to produce an output of shape $112\times112\times 32$. The rest of the convolution layers, Conv\_i, where \textit{i = 2, 3 . . 5} perform $1\times1$ convolutions to compute a linear combination of the previous output. A $1\times1$ convolution layer produces the desired number of output channels equal to the number of filters used. Batch normalization \cite{batch} applied after each layer produced better results whilst also accelerating training. We used ReLU nonlinearities as activations for all layers.

\item \textbf{DWS\_1} is modified such that Conv\_5 layer performs $3\times3$ convolutions and that is why we call it \textbf{DWS\_3}. Increasing the kernel size to $3\times3$ in the Conv\_5 layer helps in incorporating the neighborhood information in order to learn more complex features. However, it also increases the inference time on-device.

\item We replaced all the four depthwise separable convolution layers in \textbf{DWS\_1} with standard convolution layers keeping the number of filters the same. The kernel size is kept $3\times 3$ in all the layers. We call this model \textbf{FCONV\_3}. Changing the kernel size to $5\times5$, we end up with \textbf{FCONV\_5}.


\end{itemize}

\section{Experiments}
\subsection{Dataset}

All the images used in training and testing are in the JPEG format. We prepared our dataset for synthetic images, which we call \textbf{SocialMedia} dataset. We scraped seasonal greeting images and memes from various web sources. SocialMedia consists of 40K images out of which 10K images were used for testing. Our training set for synthetic images consists of around 34,994 images, including 4,994 images from Reddit Memes Dataset \footnote{https://www.kaggle.com/sayangoswami/reddit-memes-dataset/metadata}. For further evaluation for our solution, we used TextRecognitionDataGenerator \footnote{https://github.com/Belval/TextRecognitionDataGenerator} to add artificial text on camera images to make another test set for synthetic images having 30K instances. It is referenced as TRDG in Table1.\par
For natural image dataset, we picked random 50 images (if found) from each of the 602 classes from the Open Images Dataset V5 \footnote{https://storage.googleapis.com/openimages/web/download.html}, 15,620 images from Indoor Scene Recoginition Database proposed by A. Quattoni et al. \cite{indoor} which has 67 indoor categories. We also included 1,000 Motion and out of focus camera images from Blur Detection dataset \cite{blur}. To incorporate camera images with scenic text, we included 1,555 images from Total-Text \cite{totaltext} dataset proposed by CK Chng et al. The total size of our training set for camera images is around 43K. A natural scene may have different lighting conditions, various objects with contrast backgrounds, etc. To cover most of these aspects, we tested on various camera image datasets. These include test set from MIT Places365-Standard \cite{places} comprising of 384K images, 41K images from MS-COCO \cite{coco} 2014 validation set and 41K images from MS-COCO 2017 test set. Photographic Image dataset proposed by Tokuda et al. \cite{pg} having 4,850 images was also evaluated.

\subsection{Experimental Setup}
\begin{itemize}
\item We implemented the proposed architectures in the Tensorflow framework and all of the experiments were conducted on a GeForce GTX 1080ti GPU.
\item The loss function used is categorical cross-entropy. We have used Adam \cite{adam} optimizer with the decay of 1e-6. The initial learning rate was set to 0.001. 
\item We have used 5 fold cross-validation, where we randomly select one-fifth portion of the dataset for validation and train the model on the remaining. Batch size was kept 128 and we ran 200 epochs for training.
\end{itemize}

\subsection{Evaluation}
We have listed the performance of our architectures on various datasets discussed above in Table \RomanNumeralCaps{2}. All images have quality factor of 0.95. We see that DWS\_1 outperforms the other proposed models. The first convolution layer with kernel dimensions $3\times3\times32$ is common for all the models. It learns simple features based on edges like orientation and sharpness. Convolution layers in FCONV3 and FCONV5 models learn more complex features based on image content compared to the depthwise separable convolution layers in DWS\_1 and DWS\_3. These features become vital in tasks like objection recognition but may lead to misclassification in our scenario. This is evident in the results that we have obtained.
\par For testing the practical utility of the model, we evaluated DWS\_1 with different JPEG quality factors as listed in Table \RomanNumeralCaps{3}. The extent of noise introduced increases with decreasing quality factors. Edges become less sharp and images tend to lose finer details. Hence, we can expect growth in classification accuracy in case of camera images. And, this shall adversely affect accuracy in case of synthetic images. This underlying hypothesis is well supported by the results obtained in Table \RomanNumeralCaps{3}. The low running time of the model is important for a real-time experience. Fig.3 shows the plot for Inference time versus the Resolution factors of the input image. Resolution factor of \textit{r} has image size of $224r\times224r$.

We have also compared our approach to traditional methods \cite{7params,npic2}, where classification is based on hand-crafted features. For this purpose, we prepared a dataset containing 10K images for train and 2K images for test equally distributed between both the classes. Following \cite{7params}, we computed ten features per image for classification. Based on colour, different colour ratio and saturation average are used. To incorporate edge information, space correlation amongst pixels was measured in terms of features (average) like spatial gray level dependence\cite{20}, color correlogram\cite{17} (for each color channel), gray histogram \cite{7params} and farthest neighbor \cite{7params}. Table \RomanNumeralCaps{1} shows the performance of various classifiers on these hand-crafted features.

\vspace{0.4cm}
\captionof{table}{}
\begin{center}
\begin{tabular}{|c|c|}
\hline 
 \rule{0pt}{10pt}\textbf{Classifiers} & \rule{0pt}{10pt}\textbf{Test Accuracy} \\
\hline 
 \rule{0pt}{10pt}SVM & \rule{0pt}{10pt}61.21\% \\
\hline
 \rule{0pt}{10pt}Neural Network & \rule{0pt}{10pt}61.79\% \\
\hline
 \rule{0pt}{10pt}Random Forest & \rule{0pt}{10pt}\textbf{65.23}\% \\
\hline
\end{tabular}
\end{center}
\vspace{0.4cm}
	
From the above table, we see that the traditional approach does not perform well on typical social media images shared over the internet today. These failure cases are shown in Fig. 4. The presence of small artificial text on the camera image background does not add much to the value of features based on edges sharpness. Therefore, it is prone to misclassification. Moreover, with the advanced camera sensors, camera images may have high average gradient magnitudes that can result in classifiers detecting them as synthetic.

\begin{figure}[t]
\includegraphics[scale=0.55]{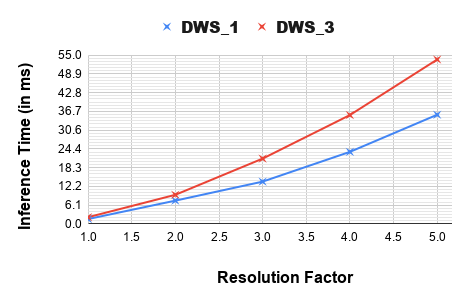}
\caption{Plot for Inference Time versus Input size}
\end{figure}

\begin{figure}[ht]
\centering
    \begin{subfigure}{\linewidth}
    \centering
        \includegraphics[width=.97\linewidth]{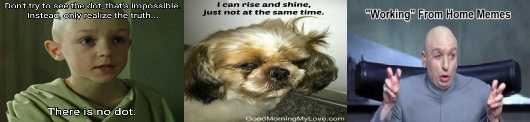}
        \caption{Synthetic Images}
    \end{subfigure}
  \\
    \vspace{0.2cm}
    \begin{subfigure}{\linewidth}
    \centering
        \includegraphics[width=.97\linewidth]{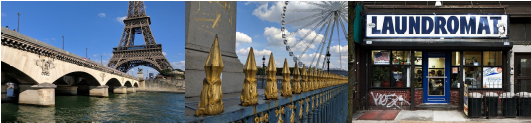}
        \caption{Camera Images}
    \end{subfigure}
    \caption{Failure Cases of Traditional Approaches}
\end{figure}

\begin{table*}[htpb]
\caption{Accuracy obtained on various public datasets}
  \label{sample-table}
  \begin{center}
  \begin{tabular}{cccccccc}
    \toprule
    \multirow{2}{*}{Models} &
    \multirow{2}{*}{No. of parameters} &
      \multicolumn{6}{c} {Datasets} \\
      & & {Places-365} & {Tokuda et al.} & {COCO-val14} & {COCO-test17} & {SocialMedia} & {TRDG} \\
      \midrule
    DWS\_1 & 67K & 98.25\% & \textbf{97.59\%} & \textbf{ 96.39\%} & \textbf{96.52\%} & \textbf{96.23\%} & \textbf{95.81\%} \\
    DWS\_3 & 328K & \textbf{98.73\%} & 96.72\% & 95.14\% & 95.02\% & 95.53\% & 95.49\%  \\
    FCONV3 & 537K & 93.47\% & 95.61\% & 93.53\% & 94.47\% & 93.27\% & 93.18\%  \\
    FCONV5 & 1,488K & 97.52\% & 94.24\% & 95.12\% & 95.04\% & 91.35\% & 93.89\% \\
    \bottomrule
  \end{tabular}
\end{center}
\end{table*}

\begin{table*}[htpb]
\caption{Performance of DWS\_1 on different Quality factors}
  \label{sample-table}
  \begin{center}
  \begin{tabular}{ccccccc}
    \toprule
    \multirow{2}{*}{Quality factors} &
      \multicolumn{6}{c} {Datasets} \\
       & {Places-365} & {Tokuda et al.} & {COCO-val14} & {COCO-test17} & {SocialMedia} & {TRDG} \\
      \midrule
    0.45 & 98.30\% & \textbf{98.39}\% & \textbf{98.22}\% & \textbf{98.31}\% & 94.45\% & 93.52\%  \\
    0.55 & 97.81\% & 98.27\% & 98.09\% & 98.29\% & 94.49\% & 93.54\%  \\
    0.65 & 98.43\% & 98.28\% & 98.17\% & 98.17\% & 94.67\% & 93.82\%  \\
    0.75 & \textbf{98.63}\% & 97.71\% & 98.18\% & 98.11\% & 95.09\% & 94.38\% \\
    0.85 & 98.28\% & 97.28\% & 98.22\% & 98.27\% & \textbf{95.37}\% & \textbf{94.79}\% \\
    \bottomrule
  \end{tabular}
\end{center}
\end{table*}

\begin{figure*}[t!]
    \begin{subfigure}[b]{0.52\textwidth}
        \includegraphics[width=\textwidth]{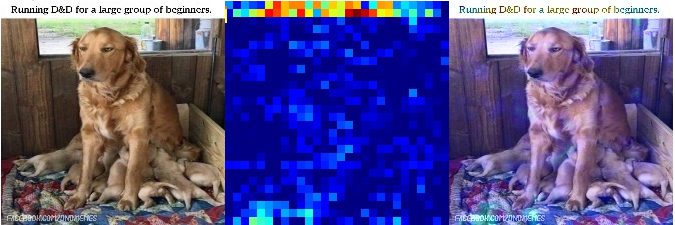}
        \caption{}
    \end{subfigure}
    \hspace*{\fill}
    \begin{subfigure}[b]{0.52\textwidth}
        \includegraphics[width=\textwidth]{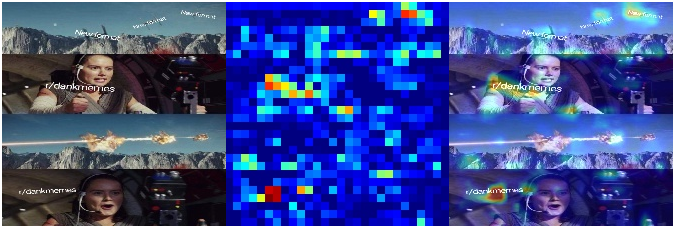}
        \caption{}
    \end{subfigure}
	\hspace*{\fill} \\
	\begin{subfigure}[b]{0.52\textwidth}
        \includegraphics[width=\textwidth]{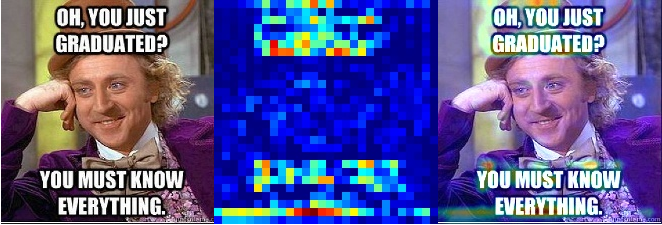}
        \caption{}
    \end{subfigure}
    \hspace*{\fill}
	\begin{subfigure}[b]{0.52\textwidth}
        \includegraphics[width=\textwidth]{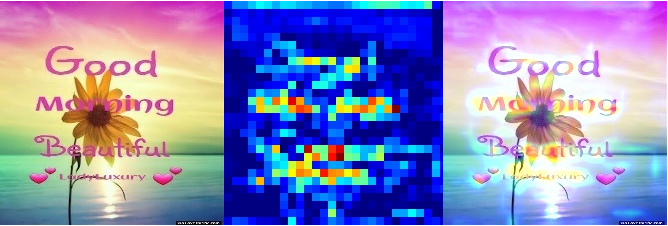}
        \caption{}
    \end{subfigure}
    \caption{Synthetic images with their heat maps}
\end{figure*}

\begin{figure*}[t!]
    \begin{subfigure}[b]{0.52\textwidth}
        \includegraphics[width=\textwidth]{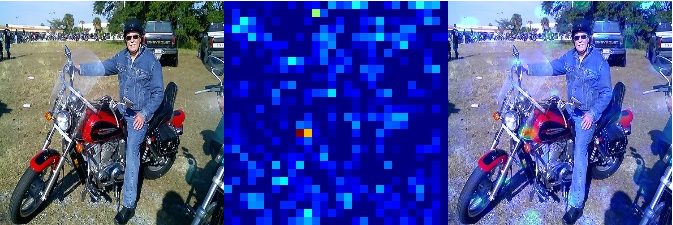}
        \caption{}
    \end{subfigure}
    \hspace*{\fill}
    \begin{subfigure}[b]{0.52\textwidth}
        \includegraphics[width=\textwidth]{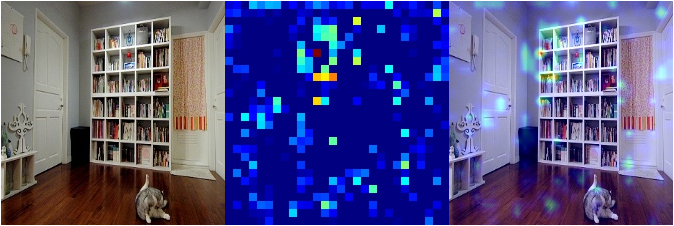}
        \caption{}
    \end{subfigure}
	\hspace*{\fill} \\
	\begin{subfigure}[b]{0.52\textwidth}
        \includegraphics[width=\textwidth]{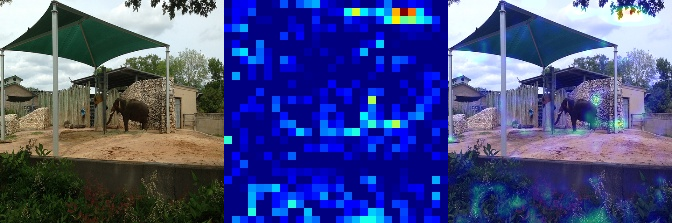}
        \caption{}
    \end{subfigure}
    \hspace*{\fill}
	\begin{subfigure}[b]{0.52\textwidth}
        \includegraphics[width=\textwidth]{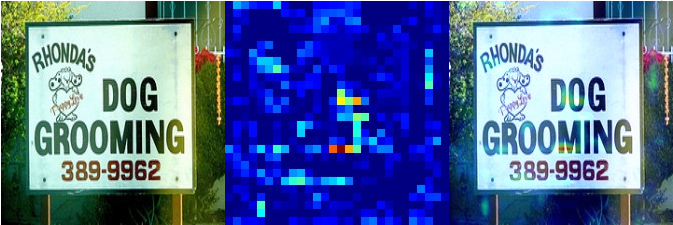}
        \caption{}
    \end{subfigure}
    \caption{Camera images with their heat maps}\label{fig:heatsyn}
\end{figure*}

\subsection{Results}
\hspace{0.2cm} Fig.5 and Fig.6 display test images along with their heat maps. These heat maps are generated from the output of Conv\_5 layer of DWS\_1 model to visualize its learning. The intensity of added artificial characteristics ranges from dark blue (lowest) to red (highest). Heat maps of synthetic images in Fig.3 have connected patches of very high intensity collectively representing the region of artificial nature whereas the heat maps of camera images shown in Fig.5 have sparsely distributed blocks (mostly single blocks) of high intensities. We have presented four different types of social media images.

\begin{itemize}
\item A thin white strip with small text is added on top of the camera image as shown in Fig.5 (a). This region is completely identified and marked with high density pixels.
\item Four different camera images having very small text being stacked together to form a meme 
as shown in Fig.5 (b). The horizontal sharp edge so formed after stacking is identified by the horizontal line of higher intensity pixels.
\item Text added on the camera image being well demarcated from the background in the heat map as shown in Fig.5 (c).
\item Fig.5 (d) shows a complete synthetic image. We can see the regions of connected high density pixels.

\end{itemize}
Various scenes are presented in Fig.6. In (d), the scenic text is shown in a contrasting white background. There are only small structures of higher intensity blocks seen over the text area and the model correctly predicts the image with confidence of 95.86\%. The solution was applied to a real scenario in a smartphone environment. We invited volunteers to test the solution. The solution was tested on around 51 users belonging to various demographics. A smart assistant interface was used to notify the users about the need for reviewing the stored or received images of various resolutions from social media. Our solution rans on the Samsung Galaxy S10 device with an average inference time of 24.3 milliseconds per image. In a real-world setting, it has been observed that the proposed approach has enriched users' efficacy in managing the device storage. We noted an average of 18\% extra available storage space (profiled weekly) for each user.

\section{Conclusion}
In this paper, we demonstrate the use of convolutional neural networks for solving the highly relevant problem of filtering social media messages. This work is the first of its kind in this problem domain and we achieved an average accuracy of 97.18\% on camera images and 96.02\% on synthetic images. We show that depthwise separable convolutions perform well in learning features to distinguish artificial features added in an image. The model shows promising results on various standard camera image datasets. Additionally, we have tested the relevance of the approach in practical scenarios. The high accuracy of the model ensures that we do not recommend a user to delete a photo that should be archived. We would like to extend this work for distinguishing photo-realistic computer graphics where edge transitions are smoother than other synthetic images on social media.


\section{References}
\begingroup
\renewcommand{\section}[2]{}%
\bibliographystyle{IEEEtran}
\bibliography{IEEEabrv,references}
\endgroup

\end{document}